\pdfoutput=1

\documentclass[11pt]{article}

\usepackage[final]{ACL2023}

\usepackage{times}
\usepackage{latexsym}

\usepackage[T1]{fontenc}

\usepackage[utf8]{inputenc}

\usepackage{microtype}

\usepackage{inconsolata}
\usepackage{booktabs}
\usepackage{makecell}
\usepackage{graphicx}
\usepackage[linesnumbered,lined,ruled,commentsnumbered]{algorithm2e}
\usepackage{amsmath}
\usepackage{amsfonts}
\usepackage{subcaption}
\usepackage{hyperref}

\usepackage{bbding}
\title{TREA: Tree-structure Reasoning Schema for Conversational Recommendation}

\author{
Wendi Li\textsuperscript{1,2}, Wei Wei\textsuperscript{1,2,\Envelope}, Xiaoye Qu\textsuperscript{1},\\
\textbf{Xianling Mao}\textsuperscript{\textbf{3}}, \textbf{Ye Yuan}\textsuperscript{\textbf{4}},\textbf{ Wenfeng Xie}\textsuperscript{\textbf{4}}, \textbf{Dangyang Chen}\textsuperscript{\textbf{4}}\\
\textsuperscript{1}Cognitive Computing and Intelligent Information Processing (CCIIP) Laboratory,\\
Huazhong University of Science and Technology\\
\textsuperscript{2}Joint Laboratory of HUST and Pingan Property \& Casualty Research (HPL)\\
\textsuperscript{3}Department of Computer Science and Technology, Beijing Institute of Technology\\
\textsuperscript{4}Ping An Property \& Casualty Insurance company of China\\
\textsuperscript{1}\texttt{\{wendili,weiw,xiaoye\}@hust.edu.cn}
\textsuperscript{3}\texttt{maoxl@bit.edu.cn}\\
\textsuperscript{4}\texttt{\{yuanye503,xiewenfeng801,chendangyang273\}@pingan.com.cn}
}

\begin{document}
\maketitle
\begin{abstract}
    \let\thefootnote\relax\footnotetext{\Envelope ~ Corresponding Author}
    Conversational recommender systems (CRS) aim to timely trace the dynamic interests of users through dialogues and generate relevant responses for item recommendations.
    Recently, various external knowledge bases (especially knowledge graphs) are incorporated into CRS to enhance the understanding of conversation contexts.
    However, recent reasoning-based models heavily rely on simplified structures such as linear structures or fixed-hierarchical structures for causality reasoning, hence they cannot fully figure out sophisticated relationships among utterances with external knowledge. To address this, we propose a novel \textbf{T}ree-structure \textbf{R}easoning sch\textbf{E}m\textbf{A} named \textbf{TREA}. TREA constructs a multi-hierarchical scalable tree as the reasoning structure to clarify the causal relationships between mentioned entities, and fully utilizes historical conversations to generate more reasonable and suitable responses for recommended results. Extensive experiments on two public CRS datasets have demonstrated the effectiveness of our approach. Our code is available at \url{https://github.com/WindyLee0822/TREA}

\end{abstract}
\section{Introduction}

Conversation Recommender System (CRS) has become increasingly popular as its superiority in timely discovering user dynamic preferences in practice.
As opposed to traditional passive-mode recommendation systems, it highlights the importance of proactively clarifying and tracing user interests through live conversation interactions, which notably enhance the success rate of item recommendations. 

\begin{figure*}[h]
  \centering
  \includegraphics[width=\linewidth]{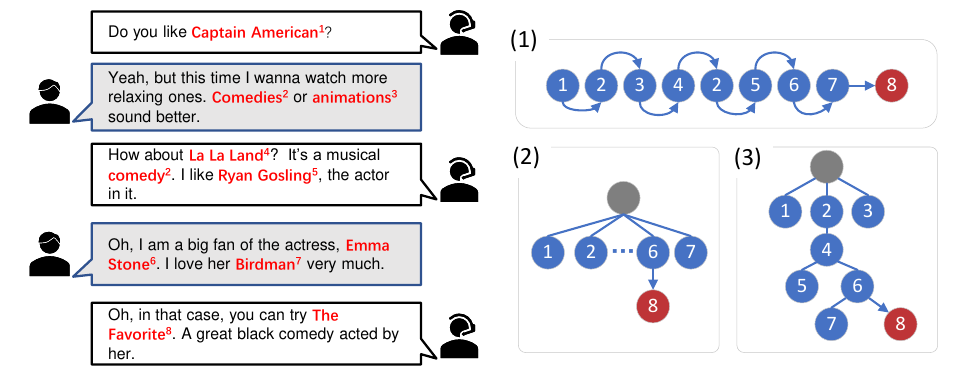}
  \caption{An example of conversational recommendation scenarios and three kinds of reasoning structures for CRS. In the conversation example, entities are marked in red and the upper-left number corresponds to the figure in the reasoning structure. (1) corresponds to the linear structure.  (2) corresponds to the structure with two fixed hierarchies (history-prediction), flattening all the mentioned entities at the first hierarchy. (3) corresponds to our multi-hierarchical structure of TREA.}
  \label{intro}
\end{figure*}

Since sole contextual utterances are insufficient for comprehensively understanding user preferences, there are many efforts devoted to incorporating various external knowledge \cite{kbrd,kgsf,c2crs,unicrs,mese}, which typically enrich the contextual information with mentioned entities recognized over utterances. 
However, these methods fail to model the complex causal relations among mentioned entities, owing to the diversity of user interest expression and the frequent shift of conversation topic as shown in Figure \ref{intro}.

Actually, it is non-trivial to explicitly model the complex causal relationships of conversations.
Although there are several reasoning-based methods proposed for CRS, their simplified structures make the objective unattainable.
Some researches \cite{crfr} track the mentioned entities as linear sequential fragments analogous to (1) in Figure \ref{intro}.
However, the linear structure is only suitable for adjacent relation modeling, which may not always work well since the actual causality between mentioned entities exists multi-hop jumps (e.g. "comedy"-"La La Land" in Figure \ref{intro}). Other studies \cite{crwalker} propose other forms of specially-designed structures for reasoning akin to (2) in Figure \ref{intro}, but they generally have fixed hierarchies, which often degenerate into a simple 2-layer hierarchy "history"-"prediction", neglecting the causal relations of historical entities. Therefore,  neither of them is applicable for full modeling of the complex reasoning causality within conversations.

To improve the reasoning capability of CRS, the challenges are twofold. The first challenge lies in empowering the model to illuminate the causal inference between all mentioned entities. To tackle this, we performs abductive reasoning for each mentioned entity to construct the multi-hierarchical reasoning tree. 
The reasoning tree explicitly preserves logical relations between all entities and can be continuously expanded as the conversation continues, which provides the model with a clear reference to historical information for prediction.
 The second challenge is how to utilize reasoning information in response generation. We enable the model to extract relevant textual information from the historical conversation with the corresponding reasoning branch, thus promoting the correlation between generated responses and recommended items. We name this \textbf{T}ree-structure \textbf{R}easoning sch\textbf{E}m\textbf{A} \textbf{TREA}.

To validate the effectiveness of our approach, we conduct experiments on two public CRS datasets. Experimental results show that our TREA outperforms competitive baselines on both the recommendation and conversation tasks. Our main contributions are summarized as follows:
\begin{itemize}
    \item To the best of our knowledge, it is the first trial of CRS to reason every mentioned entity for its causation.
    \item We propose a novel tree-structured reasoning schema to clarify the causality relationships between entities and mutual the reasoning information with the generation module.  
    \item Extensive experiments demonstrate the effectiveness of our approach in both the recommendation and conversation tasks.

\end{itemize}
\section{Related Work}
Conversational Recommender System (CRS) explores user preference through natural language dialogues. Previous works can be roughly categorized into two types. The first category of CRS is recommendation-biased CRS \cite{CRM,CPR,EAR,UNICORN,mimc}. This category focuses solely on interactive recommendations but the function of natural language is ignored. Several fixed response templates are preset on the agents and users cannot use free text but only have limited options, which can be detrimental to the user experience. 

The other category of CRSs is dialog-biased CRS \cite{redial,a1,a2,a3,a4}. This category emphasizes the critical effect of natural language, aiming to understand user utterances for accurate recommendations and generate human-like responses. 
Noticing that entities \cite{gu2022delving,qu2022distantly,qu2023survey} mentioned in conversations are important cues for modeling user preferences, \citet{kbrd} firstly integrates KG to enhance the user representation. \citet{kgsf,ntrd} use two KGs on entity-granularity and word-granularity respectively to represent the user preference more comprehensively. Subsequent researches introduce other types of external knowledge e.g. item description \cite{revcore,c2crs} or pretrained language models (PLMs) \cite{mese,unicrs} to further assist the user representations. However, they commonly treat each mentioned knowledge piece equally and integrate them into an aggregated representation. 

\begin{figure*}[t]
  \centering
  \includegraphics[width=\linewidth]{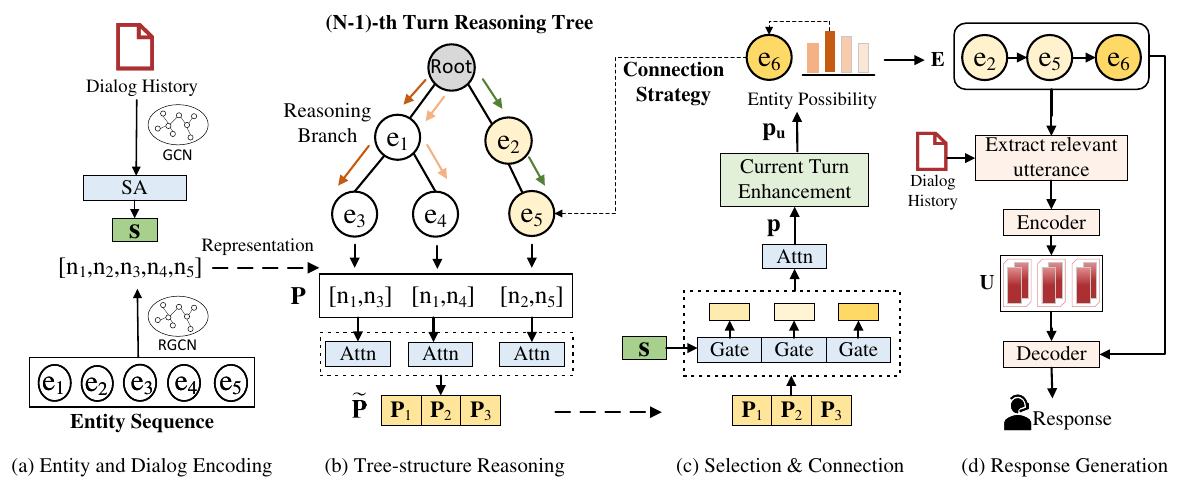}
  \caption{The overview of our proposed TREA. We first encode the entities and the sentences in the dialog history. Then we aggregate the information of each reasoning branch in the current reasoning tree. Later, a comprehensive representation of dialog semantics measures the devotion of each reasoning branch to the current recommendation. After the current turn enhancement, we select the entity to join the reasoning tree with the connection strategy. The extended reasoning branch guide the extraction of relevant textual information for the generation module.
  }
  \label{modelfig}
\end{figure*}

Recently, some researches manage to model the reasoning process during conversations. \citet{crfr} linearize the mentioned entity sequence and reasoning the inferential causality between the adjacent entity pairs. \citet{crwalker} create non-linear reasoning structures, but they do not preserve the hierarchy of historical turns. 
Therefore these reasoning methods have limited performance improvement.

To sort out the causal relations among utterances, our model performs tree-structured reasoning on the entire dialogue history for each mentioned entity. We also inject the reasoning information into the generation process to make responses more relevant, achieving that the reasoning process facilitates both recommendation and generation tasks simultaneously.




\section{Methods}

In this section, we present the Tree-structure reasoning schema TREA as demonstrated in Figure 
\ref{modelfig}. Specifically, we first introduce the encoding of entities and word tokens. Then we illustrate the construction procedure of the reasoning tree. Later, we describe how the reasoning information supports the generation module. Finally, we explain the process of parameter optimization.

\subsection{Entity and Dialog Encoding}
Following previous works \cite{kbrd,kgsf,crwalker,c2crs}, we first perform entity linking based on an external KG DBpedia \cite{dbpedia}, and then encode the relational semantics via a relational graph neural network (RGCN) \cite{rgcn} to obtain the corresponding entity embeddings. Formally, the embedding $\textbf{n}_e^{l+1}$ of entity $e$ at the $l+1$-th graph layer is calculated as:
\begin{equation}
\textbf{n}_{e}^{l+1} = \sigma(\sum_{r\in \mathcal{R}} \sum_{e'\in \mathcal{N}_e^r} \frac{1}{Z_{e,r}} \textbf{W}_r^l \textbf{n}_{e'}^l + \textbf{W}^l \textbf{n}_e^l) \label{eqrgcn}
\end{equation}
where $R$ is a relation set, $\mathcal{N}_e^r$ denotes the set of neighboring nodes for $e$ under the relation $r$, $\textbf{W}_r^l,\textbf{W}^l$ are learnable matrices for relation-specific aggregation with neighboring nodes and representation transformation respectively, $Z_{e,r}$ is a normalization factor, $\sigma$ denotes the sigmoid function. The semantic information of word tokens is encoded by an external lexical knowledge graph ConceptNet \cite{conceptnet}. We further adopt a graph convolutional neural network (GCN) \cite{kipf2016semi} to propagate and aggregate information over the entire graph.

\subsection{Reasoning Tree Construction.}
The construction of reasoning trees is introduced in a manner similar to mathematical induction. We first explain the structure initialization at the first conversation round, then illustrate the structure transition from the $(n$-$1)$-th round to the $n$-th round. The structure of the whole tree can be deduced accordingly.

To initialize the reasoning tree, we first set a pseudo node as the root node. The root node does not represent any entity in the conversations but is just a placeholder. When the first utterance is coming, the first mentioned entity is directly connected to the root node. The subsequent entities in the first utterance are connected following the Algorithm \ref{connect}.

When the conversation progresses to $(n$-$1)$-th round, the known conditions are as follows: the current reasoning tree $\mathcal{T}_{n-1}$, utterance tokens sequences $s_t$. They are utilized for the extension of the reasoning tree $\mathcal{T}_{n-1}$, which is described in two parts, tree-structure reasoning and the selection \& connection of candidate entities.

\textbf{Tree-Structure Reasoning.} We embed all the reasoning branches and pad them to a certain length $l_r$. A path from the root node to any leaf node of the tree is referred to as a \textit{reasoning branch} since it expresses a chain of coherent inferences. To represent the sequential information for each reasoning branch, we inject a learnable position embedding into the embedding of each entity element. The position-enhanced branch embedding matrix is denoted as $\textbf{P} \in \mathbb{R}^{n_r\times l_r\times d}$ where $n_r$ is the branch number of $\mathcal{T}_{n-1}$ and $d$ is the dimension of embeddings. We incorporate a linear attention mechanism to integrate the representation of each path. The attention scores are calculated as follows:
\begin{equation}
\begin{aligned}
    \widetilde{\textbf{P}} &= \textrm{Attn}(\textbf{P}) = \textbf{P} \, \mathbf{\alpha}_r \\
    \mathbf{\alpha}_r &= \textrm{Softmax}(\mathbf{b}_r \tanh (\textbf{W}_r \textbf{P})) 
    \label{selfattn}
\end{aligned}
\end{equation}
where $\textbf{W}_r,\mathbf{b}_r$ are learnable parameters.
Embeddings of entities in a certain reasoning branch are aggregated according to the attention score. Then we can obtain the comprehensive representations of reasoning branches denoted as $\widetilde{\textbf{P}} \in \mathbb{R}^{n_r\times d}$.

\textbf{Selection \& Connection.} Since the reasoning branches have varying-degrees contributions to the next-hop entity, the model analyzes the semantics of word tokens $s_t$ to measure the impact of each branch. The formulas are as follows:
\begin{equation}
\begin{aligned}
    \textbf{p} &=\textrm{Attn}(\gamma \widetilde{\textbf{P}} + (1-\gamma)\mathbf{s} ) \\
    \gamma &= \sigma ( \textbf{W}_s \textrm{Concat}(\widetilde{\textbf{P}} \,, \mathbf{s} ))
\label{gate}
\end{aligned}
\end{equation}
where $\mathbf{W}_s$ is a learnable parameter, $\mathbf{s}$ is the comprehensive semantic representation of the word tokens in ConceptNet which are aggregated with the linear attention mechanism in Eq.\ref{selfattn}. Then we can obtain the user representation $\textbf{p}_u$ that combines semantic and reasoning information.
Since the latest turn has a prominent significance to the response \cite{uccr}, we collect the entities and word tokens from the current conversation turn, embedded to $\textbf{e}_c,\textbf{s}_c$. Then we aggregate the current turn information and mutual it with acquired representation $\textbf{p}$ as follows:
\begin{equation}
    \textbf{p}_u = g(\textbf{p}, g'(\textrm{Attn} (\textbf{e}_c),\textrm{Attn}(\textbf{s}_c))
\label{user}
\end{equation}
where $g(\cdot\,,\cdot),g'(\cdot\,,\cdot)$ are two gate layers like Eq.\ref{gate}. Then we derive the next-hop possibility distribution from the overall user representation:
\begin{equation}
    \mathcal{P}_r^u = \textrm{Softmax}([\textbf{p}_u \textbf{e}_0^\textrm{T},\cdots,\textbf{p}_u \textbf{e}_n^\textrm{T}])
\end{equation}
where $\textbf{e}_0,\cdots,\textbf{e}_n$ are representations of all entities. The entity with the largest probability is selected and connected to the reasoning tree. The connection strategy is as Algorithm \ref{connect}.
\begin{algorithm}
\SetKwData{e}{$e^*$}\SetKwData{ES}{$ES$}
\SetKwFunction{isadj}{IsAdj}
\SetKwFunction{addedge}{AddEdge}
\SetKwInOut{Input}{input}
\Input{Selected entity $e^*$; Entity sequence $ES$ in reverse order of mention; Reasoning Tree $\mathcal{T}$ with  root node $r$}
\ForEach {$e$ in $ES$}
{\If{\isadj{$e$,\e}} {// \emph{Two entities are adjacent in KG}\;
\addedge{$e$,\e}\; // \emph{Add an edge $(e,e^*)$ in $\mathcal{T}$}\;
return } 
}
\addedge{$r$,\e}\;
return
\caption{Connection Strategy}
\label{connect}
\end{algorithm}

\subsection{Reasoning-guided Response Generation}
After adding the predicted entity to the reasoning tree, the objective of the conversation module is to generate utterances with high relevance to the predicted entity. Reasoning branches that involve the new entity 
and the historical utterances that mention the relevant entities in branches are extracted, which are encoded by RGCN and standard Transformer \cite{transformer} respectively. The corresponding embedding matrices are denoted as $\textbf{E},\textbf{U}$.
Following \cite{kgsf}, we incorporate multiple cross-attention layers in a Transformer-variant decoder to fuse the two groups of information. The probability distribution over the vocabulary is calculated as follows:
\begin{align}
    \textbf{R}^l &= \textrm{Decoder}(\textbf{R}^{l-1},\textbf{E},\textbf{U}) \label{decoder}\\ 
    \textbf{R}^b &= \textrm{FFN}(\textrm{Concat}(\textrm{Attn}(\textbf{E}),\textbf{R}^l)) \label{copy}\\
    \mathcal{P}_g&= \textrm{Softmax}(\textbf{R}^l \textbf{V}^{\textrm{T}} + \textbf{R}^b \textbf{W}^v)
\end{align}
where \textbf{V} is the embedding matrix of all words in the vocabulary, $\textbf{W}^v$ is a learnable parameter that converts the $\textbf{R}^b$ dimension to $|\textbf{V}|$. The copy mechanism is adopted in Eq.\ref{copy} to enhance the generation of knowledge-related words. The transformation chain \cite{kgsf} in the decoder of Eq.\ref{decoder} is $\textit{generated words} \!\rightarrow \!\textit{relevant entities}\!\rightarrow \!\textit{historical utterances}$.

\subsection{Optimization}

The parameters can be categorized into two parts, the reasoning parameters and the generation parameters, denoted by $\theta_r,\theta_g$. The reasoning objective is to maximize the predicted probability of the upcoming entity. The cross-entropy loss is adopted to train the reasoning module. During the training, we propose two auxiliary loss functions, isolation loss to maintain the independence of each reasoning branch, and alignment loss to bridge the representation gap. 

\textbf{Isolation Loss.} Since reasoning branches that have no shared parts are generally irrelevant, representations from different reasoning branches are expected to be dissimilar. To maintain the isolation of each reasoning branch, we propose isolation loss. Given representations of different reasoning branches, the isolation loss is calculated as
\begin{equation}
    \mathcal{L}_{I} = \sum_{i\neq j} \textrm{sim}(\widetilde{\textbf{p}}_i, \widetilde{\textbf{p}}_j) = \sum_{i \neq j} \frac{\widetilde{\textbf{p}}_i \widetilde{\textbf{p}}_j} {\left|\widetilde{\textbf{p}}_i\right| \cdot \left|\widetilde{\textbf{p}}_j \right|}
\end{equation}
where $\widetilde{\textbf{p}}_i,\widetilde{\textbf{p}}_j$ are representations of two different reasoning branches extracted from $\widetilde{\textbf{P}}$. 

\textbf{Alignment Loss.}
The representation gap exists between the semantics and the entities since their encoding processes are based on two separate networks. Hence the entity representation and semantic representation of the same user should be dragged closer; those of different users should be pushed further to reduce the gap. The formula is as follows:
\begin{equation}
    \mathcal{L}_{a}= \lambda_c \text{sim}(\textbf{p}_c,\textbf{s}_c) + (1-\lambda_c) \textrm{sim}(\textbf{p},\textbf{s})
\label{aligneq}
\end{equation}
where $\textbf{p}_c,\textbf{s}_c$ are aggregated representation $\textrm{Attn}(\textbf{e}_c),\textrm{Attn}(\textbf{w}_c)$ in Eq.\ref{user}, $\lambda_c$ is a hyperparameter.

Then We can optimize parameters $\theta_r$ through the following formula:
\begin{equation}
    \mathcal{L}_{r}= -\sum_{u}\sum_{e_i}\log \mathcal{P}_{r}^u[e_i] + \lambda_I \mathcal{L}_{I} + \lambda_{a} \mathcal{L}_{a}
    \label{reasonloss}
\end{equation}
where $e_i$ is the order of the target entity at the $i$-th conversation round of user $u$, $\lambda_{I}, \lambda_{al}$ are hyperparameters.

When the reasoning loss $\mathcal{L}_r$ converges, we optimize the parameters in $\theta_g$. After obtaining the relevant entities and utterances via the reasoning tree, we calculate the probability distribution of the next token. To learn the generation module, we set the cross-entropy loss as:
\begin{equation}
    \mathcal{L}_g = -\frac{1}{N} \sum_{t=1}^N {\log \mathcal{P}_g^t(s_t|s_1,s_2,\ldots,s_{t-1}) }
\end{equation}
where $N$ is the number of turns in a certain conversation $C$. We compute this loss for each utterance $s_t$ from $C$.

\section{Experiment}

\subsection{Dataset.} 
We conduct our experiments on two widely-applied benchmark
datasets on CRS, which are multilingual including English (ReDial) and Chinese (TG-ReDial). \textbf{ReDial}\cite{redial} collects high-quality dialogues for recommendations on movies through crowd-sourcing workers on Amazon Mechanical Turk(AMT). The workers create conversations for the task of movie recommendation in a user-recommender pair setting following a set of detailed instructions. It contains 10,006 conversations consisting of 182,150 utterances.
\textbf{TG-ReDial}\cite{tgredial}  is annotated in a semi-automatic way. It emphasizes natural topic transitions from non-recommendation scenarios to the desired recommendation scenario. Each conversation includes a topic path to enforce natural semantic transitions. It contains 10,000 conversations consisting of 129,392 utterances.

\begin{table*}[]
    \centering
    \scalebox{0.95}{
     \setlength{\tabcolsep}{1mm}{
    \begin{tabular}{c|cccccc|cccccc}
    \toprule
        Dataset & \multicolumn{6}{|c|}{ReDial} & \multicolumn{6}{|c}{TG-ReDial} \\
    \midrule
        Method & R@10 & R@50 & Dist-3 & Dist-4 & Bleu-2 & Bleu-3 & R@10 & R@50 & Dist-3 & Dist-4 & Bleu-2 & Bleu-3   \\
    \midrule
        ReDial & 0.140 & 0.320 & 0.269 & 0.464 & 0.022 & 0.008 & 0.002 & 0.013 & 0.529 & 0.801 & 0.041 & 0.010 \\

        KBRD & 0.150 & 0.336 & 0.288 & 0.489 & 0.024 & 0.009 & 0.032 & 0.077 & 0.691 & 0.997 &  0.042 & 0.012 \\

        KGSF & 0.183 & 0.377 & 0.302 & 0.518 & 0.025 & 0.009 & 0.030 & 0.074 & 1.045 & 1.579 & 0.046 & 0.014 \\

        RevCore & 0.204 & 0.392 & 0.307 & 0.528 & 0.025 & 0.010 & 0.029 & 0.075 & 1.093 & 1.663 & 0.047 & 0.014 \\

        CR-Walker & 0.187 & 0.373 & 0.338  & 0.557 & 0.024 & 0.009 & - & - & - & - & - & - \\
        
        CRFR & 0.202 & 0.399 & 0.516 & 0.639 & - & - & - & - & - & - & - & - \\

        $\textrm{C}^2\!\textrm{-CRS}$ & 0.208 & 0.409 & 0.412 & 0.622 & 0.027 & 0.012 & 0.032 & 0.078 &  1.210 & 1.691 & 0.048 & 0.015 \\

        UCCR & 0.202 & 0.408 &  0.329 & 0.564 & 0.026 & 0.011 & 0.032 & 0.075  & 1.197 & 1.668 & 0.049 & 0.014 \\
    \midrule
        \textbf{TREA} & \textbf{0.213}$^*$ & \textbf{0.416}$^*$ & \textbf{0.692}$^*$ & \textbf{0.839}$^*$ & \textbf{0.028}$^*$ & \textbf{0.013}$^*$ &\textbf{0.037}$^*$ & \textbf{0.110}$^*$ & \textbf{1.233}$^*$ & \textbf{1.712}$^*$ & \textbf{0.050}$^*$ & \textbf{0.017}$^*$ \\
    \bottomrule
    \end{tabular}}}
    \caption{Automatic evaluation results on two datasets. Boldface indicates the best results. Significant improvements over best baseline marked with $^*$.(t-test with $p<0.05$)}
    \label{autoresults}
\end{table*}

\subsection{Baselines} 
We evaluate the effectiveness of our model with following competitive baselines:

\textit{ReDial} \cite{redial} comprises a conversation module based on hierarchical encoder-decoder architecture\cite{HERD} and a recommendation module based on auto-encoder.

\textit{KBRD} \cite{kbrd} firstly utilizes KG to enhance the user representation. The Transformer\cite{transformer} architecture is applied in the conversation module.

\textit{KGSF} \cite{kgsf} incorporate two external knowledge graphs on different aspects to further enhance the user representations. The KG information is employed in the decoding process.



\textit{CRFR} \cite{crfr} can generate several linear reasoning fragments through reinforcement learning to track the user preference shift.

\textit{CR-Walker} \cite{crwalker} create a two-hierarchy reasoning tree between history and forecast and preset several dialog intents to guide the reasoning. 

\textit{$\textrm{C}^2\!\textrm{-CRS}$} \cite{c2crs} proposed a contrastive learning based pretraining approach to bridge the semantic gap between three external knowledge bases.

\textit{UCCR} \cite{uccr} considers multi-aspect information from the current session, historical sessions, and look-alike users for comprehensive user modeling.

\subsection{Metrics} 
For recommendation evaluation, we used \textit{Recall@n} (R@n,n=10,50), which shows whether the top-n recommended items include the ground truth suggested by human recommenders.
For the response generation task, we evaluate models by \textit{Bleu-n}(n=2,3) \cite{bleu}, \textit{Dist-n}(n=3,4) \cite{distn} for word-level matches and diversity. To evaluate the generation performance more equitably, three annotators are invited to score the generated candidates from two datasets for human evaluation on the following three aspects: \textit{Fluency}, \textit{Relevance}, and \textit{Informativeness}. The inter-annotator coherence is measured by Fleiss’ Kappa.

\subsection{Implementation Details}
We keep the same data preprocessing steps and hyperparameter settings as previous researches \cite{c2crs,crwalker}. We adopt the same mask mechanism as NTRD\cite{ntrd}. The embedding dimensions of reasoning and generation are set to 300 and 128 respectively. In the encoding module, the word embeddings are initialized via Word2Vec\footnote{\url{https://radimrehurek.com/gensim/models/word2vec.html}} and the layer number is set to 1 for both GNN networks. The normalization constant of RGCN is 1. We use Adam optimizer \cite{adam} with the default parameter setting. For training, the batch size is set to 64, the learning rate is 0.001, gradient clipping restricts the gradients within [0,0.02]. For hyperparameters, $Z_e,r$ of RGCN in Eq.\ref{eqrgcn} is $1$, $\lambda_c$ of representation alignment in Eq.\ref{aligneq} is $0.9$, $\lambda_I,\lambda_{a}$ in Eq.\ref{reasonloss} is $0.008,0.002$ respectively. 

\subsection{Overall Performance Analysis}
\textbf{Recommendation.} The columns R@10,R@50 of Table \ref{autoresults} present the evaluation results on the recommendation task. It shows that our TREA significantly outperforms all the baselines by a large margin on both two datasets, which verifies that TREA can clarify the sophisticated causality between the historical entities and accurately model the user preferences. Moreover, even though RevCore and $\textrm{C}^2\!\textrm{-CRS}$ utilize the additional knowledge, they are still not as effective as TREA, which further proves the significance of correct reasoning. CR-walker and CRFR are two previous methods that manage to reason over the background knowledge. CR-Walker does not preserve the hierarchy between the historical information and CRFR linearizes the reasoning structure. Therefore even though CR-walker conducts the additional annotations of dialog intents and CRFR applies the reasoning on another KG to assist, the performance raising is limited, which certifies that our non-linear tree-structured reasoning over all mentioned entities does facilitate the user modeling.  

\begin{table}[!h]
    \centering
    \begin{tabular}{ccccc}
    \toprule
        Method & Rel. & Inf. & Flu. & Kappa \\
    \midrule
        RevCore & 1.98 & 2.22 & 1.53 & 0.78 \\
        CR-Walker & 1.79 & 2.15 & 1.68 & 0.77 \\
        $\textrm{C}^2\!\textrm{-CRS}$ & 2.02 & 2.25 & 1.69 & 0.66 \\
        UCCR & 2.01 & 2.19 & 1.72 & 0.72 \\
    \midrule
        \textbf{TREA} & \textbf{2.43} &\textbf{ 2.26} & \textbf{1.75} & 0.75 \\
    \bottomrule
    \end{tabular}
    \caption{Human evaluation results on the conversation task. Rel., Inf. and Flu. stand for Relevance, Informativeness and Fluency respectively. Boldface indicates the best results (t-test with $p<0.05$).}
    \label{humaneval}
\end{table}
\textbf{Generation.} The columns Dist-n, Bleu-n of Table \ref{autoresults} present the automatic evaluation results on the conversation task. Since CR-walker adopts GPT-2 in the original model, we initialize the generation module with Word2Vec instead for a fair comparison. It shows that TREA surpasses all baselines on generation diversity and matchness. 
Table \ref{humaneval} presents the human evaluation results. All Fleiss’s kappa values exceed 0.6, indicating crowd-sourcing annotators have reached an agreement. The results show that our TREA leads to a higher relevance of generated utterances.  It can be derived that the extraction of relevant information with the reasoning tree does improve the relevance of the generation. 

\subsection{Ablation Study}
\textbf{Recommendation.} 
\begin{figure}[t]
  \centering
  \begin{subfigure}{\linewidth}
		\centering	
        \includegraphics[width=1\linewidth]{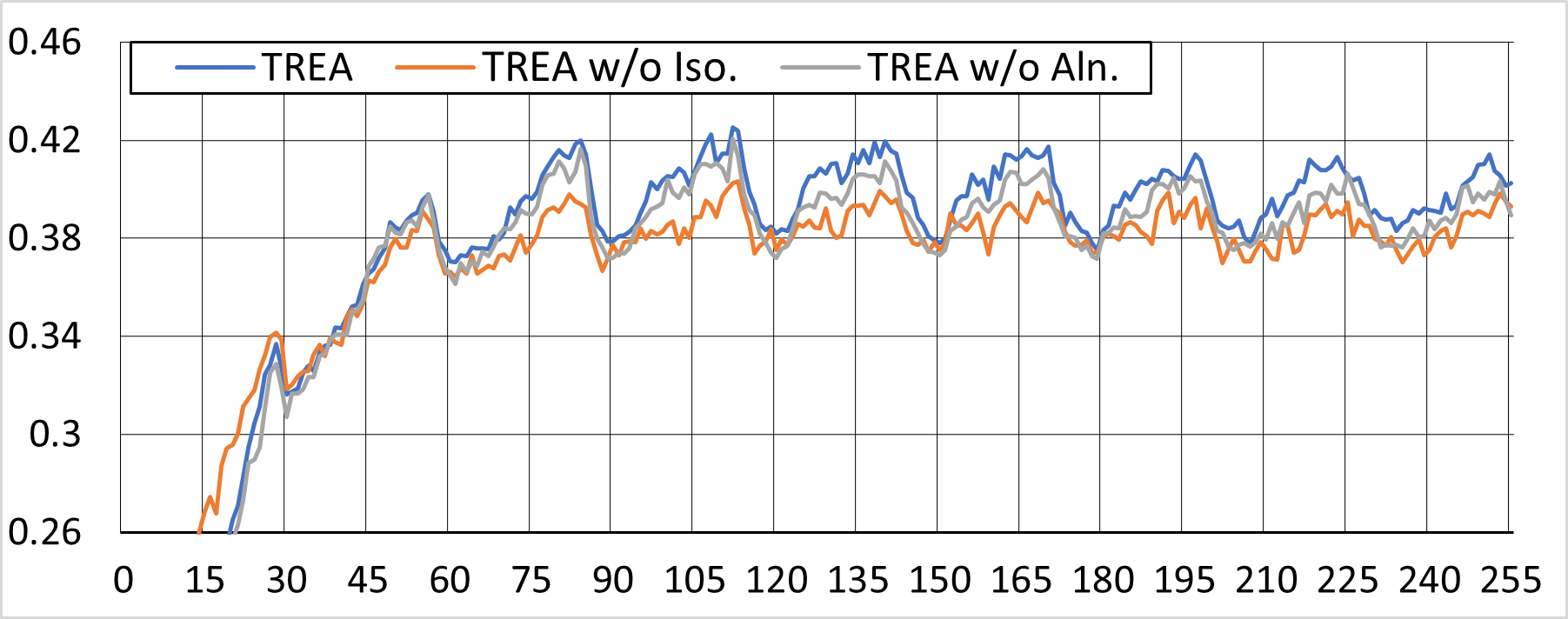}
        \caption{Recall@50}
        \end{subfigure}\\
	\begin{subfigure}{\linewidth}
		\centering
        \includegraphics[width=1\linewidth]{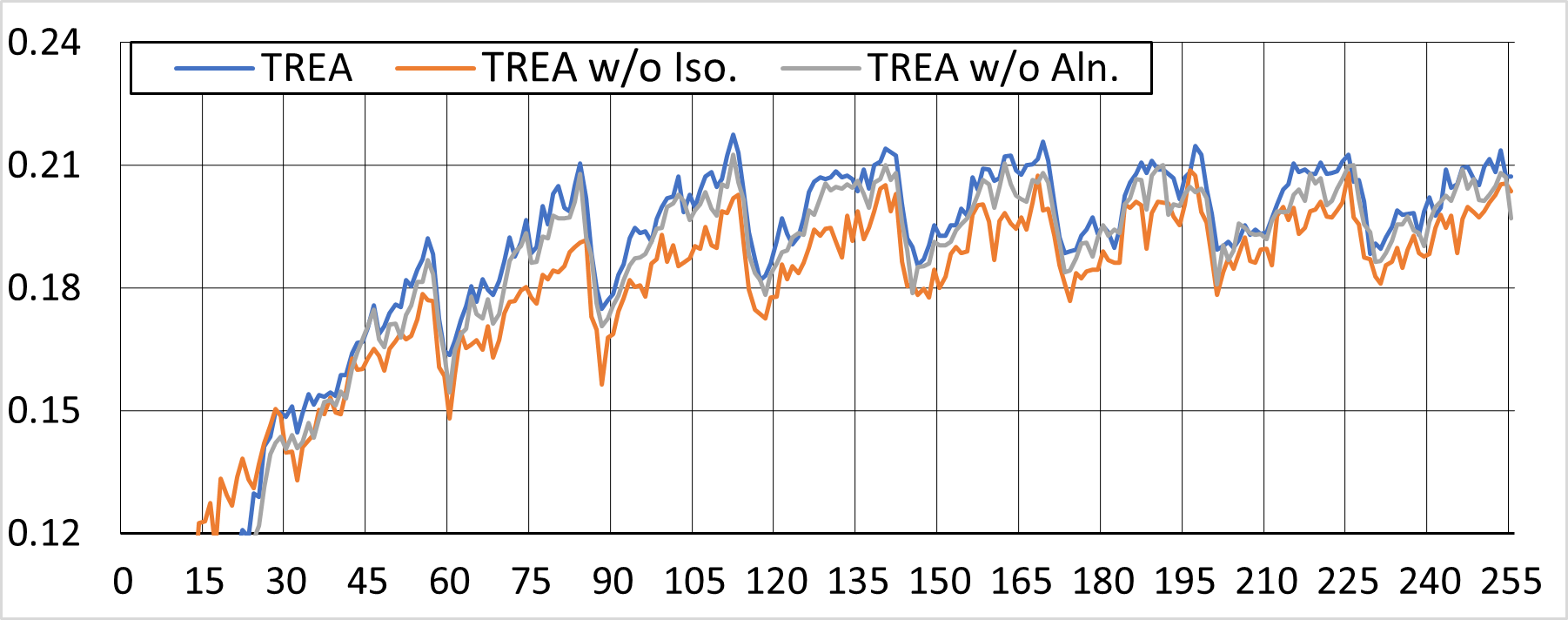}
        \caption{Recall@10}
	\end{subfigure}
  \caption{Performance comparison of TREA and its two variants. One step (X-axis) denotes parameter updates for 20 batches of training data.
  }
  \label{ablrec}
\end{figure}
\begin{table}[t]
    \centering
    \setlength{\tabcolsep}{1mm}{
    \begin{tabular}{c|cc|cc}
    \toprule
        Dataset & \multicolumn{2}{|c|}{ReDial} & \multicolumn{2}{|c}{TG-ReDial}\\
    \midrule
        Method & R@10 & R@50 & R@10 & R@50 \\
    \midrule
         TREA & 0.214 & 0.418 & 0.037 & 0.110\\
         TREA w/o Iso. & 0.202 & 0.405 & 0.028 & 0.079\\
         TREA w/o Aln. & 0.209 & 0.412 & 0.035 & 0.103 \\
         TREA w/o IA. & 0.201 & 0.403 & 0.026 & 0.076 \\
    \bottomrule
    \end{tabular}}
    \caption{Ablation results on the recommendation task. (t-test with $p<0.05$)}
    \label{ablrectable}
\end{table}
\begin{figure}[!htbp]
	\centering
	\includegraphics[width=0.48\textwidth]{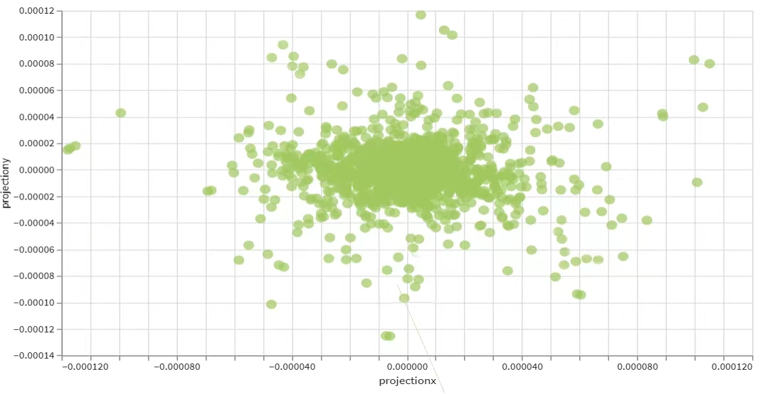} \\
	\includegraphics[width=0.48\textwidth]{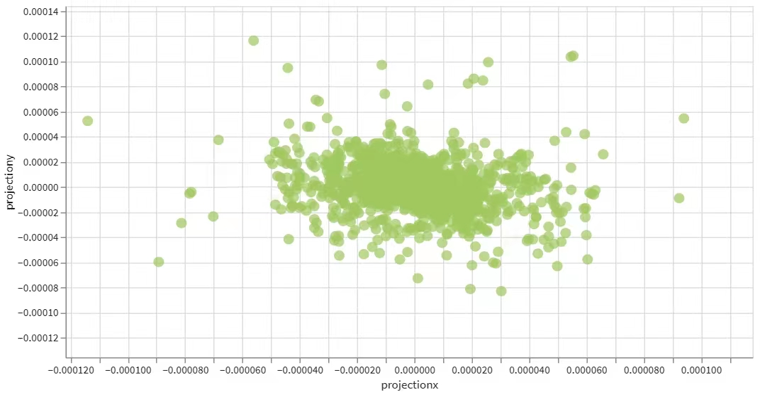}
	\caption{2D projection of KG embeddings trained by TREA (the above) and TREA w/o Iso. (the below) to illustrate the impact of the isolation loss $\mathcal{L}_I$. Embeddings are projected through t-SNE with Perplexity set to 10 and the Iterations set to 13.)}
	\label{embabl}
\end{figure}
The parameter optimization for the reasoning module involves two additional loss, isolation loss (Iso.) $\mathcal{L_I}$ and alignment loss (Aln.) $\mathcal{L}_{a}$. We would like to verify the effectiveness of each part. We incorporate three variants of our model for ablation analysis on the recommendation task, namely \textit{TREA w/o Iso.}, \textit{TREA w/o Aln.} and \textit{TREA w/o IA.}, which remove the isolation loss, the alignment loss and both of them respectively. As shown in Table \ref{ablrectable}, both components contribute to the final performance. Furthermore, we can see that removing the isolation loss leads to a large performance decrease, which suggests that maintaining the representation dependence of each reasoning branch is crucial to the correctness of the reasoning. 

To further confirm that the performance improvement is consistent and stable instead of accidental. We test the models under different iteration steps and display the corresponding results in Figure \ref{ablrec}. It can be seen that when the training loss converges, each ablation component contributes to the model performance regardless of the iteration number, which proves that the two additional loss functions are stably effective.

\textbf{The Effect of Isolation Loss.} The above subsection has verified the great impact of the isolation loss. We take a deeper dive to determine how it benefits model performance. If removing the isolation loss, since each reasoning branch participates in the calculation of the predicted possibility distribution, the representations of entities in different reasoning branches would approach each other for sharper descending of the loss value, which means that the representation of unrelevant entities would get similar irrationally and finally lead to the representation convergence of the entire knowledge graph. To confirm the assumption, we display the entity embeddings trained by TREA and TREA w/o Iso. in Figure \ref{embabl}. It shows that representations of KG entities in model without the isolation loss are more congested and less distinguishable. It demonstrates the isolation loss can prohibit the clustering of the nodes in KG, which is consistent with the above conjecture.

\textbf{Generation.} To examine whether the extraction of the relevant information through the reasoning tree benefits the generation, we conduct the ablation study based on three variants of our complete model, which utilize the whole historical entities, the whole historical utterances and both of the above without extraction, namely \textit{TREA w/o Ent.}, \textit{TREA w/o Utt.}, \textit{TREA w/o EU.} respectively. The results in Table \ref{ablgen} show that deleting either extraction brings a performance decrease on all generation metrics. PPL (Perplexity) is an automatic evaluation metric for the fluency of generations and confidence in the responses. The results of PPL show that the extraction of the relevant information reduced the model confusion. A substantial decrease on Rel. shows that reasoning-guided extraction especially influences the relevance of the generation.

\begin{table}[h]
    \centering
    \setlength{\tabcolsep}{1mm}{
    \begin{tabular}{ccccc}
    \toprule
        Model & Dist-4 & Bleu-3 & PPL\small{($\downarrow$)} & Rel. \\ 
    \midrule
        TREA & 0.839 & 0.013 & 4.49 & 2.43 \\
        TREA w/o Ent. & 0.799 & 0.012 & 4.56 & 2.28\\
        TREA w/o Utt. & 0.764 & 0.011 & 4.61 & 2.13\\
        TREA w/o EU. & 0.789 & 0.011  & 4.78 & 2.10\\
    \bottomrule
    \end{tabular}}
    \caption{Evaluation results on the ablation study of the generation task. Fleiss’s kappa values of Rel. all exceed 0.65.}
    \label{ablgen}
\end{table}
\begin{figure}[h]
	\begin{subfigure}{0.49\linewidth}
		\centering	
        \includegraphics[width=1\linewidth]{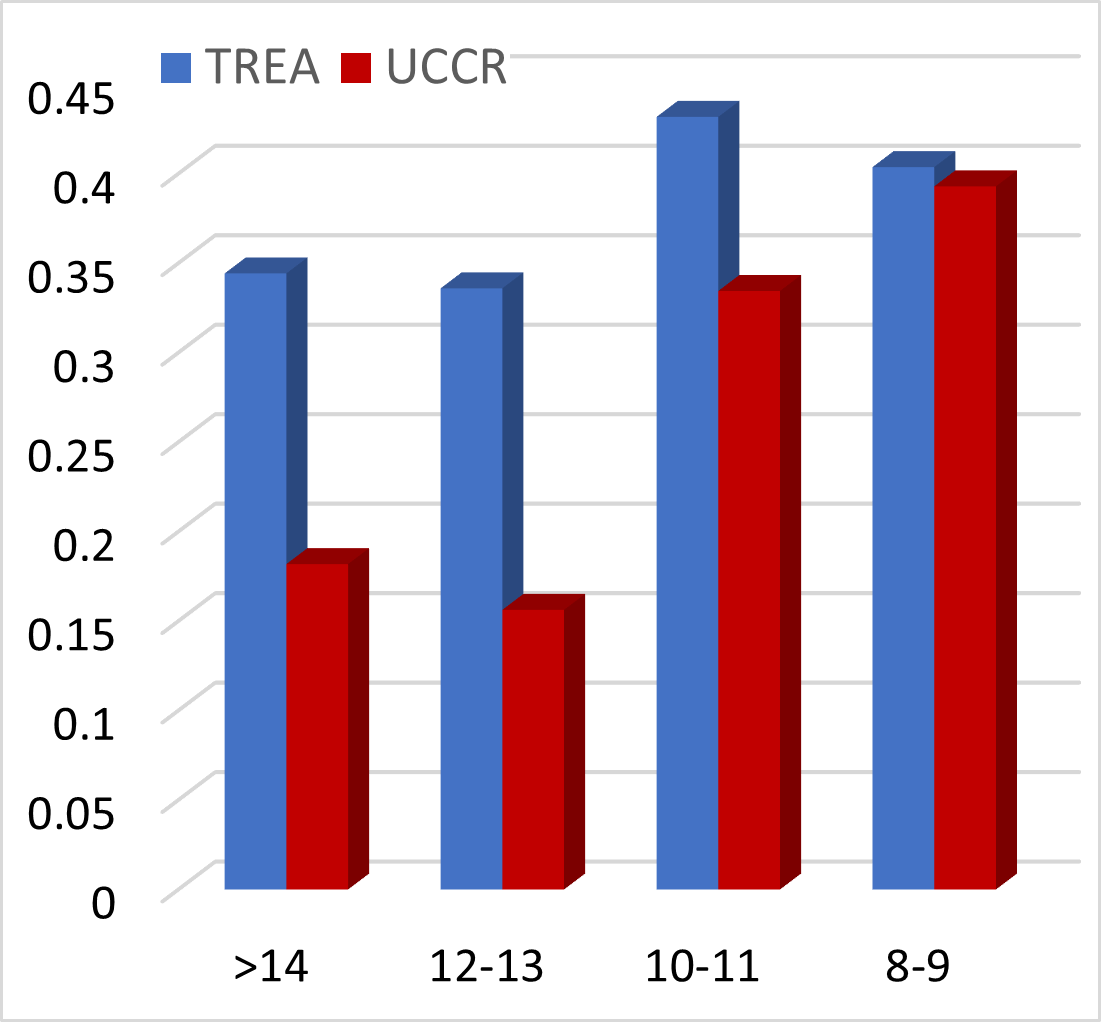}
        \caption{ReDial}
        \end{subfigure}
	\begin{subfigure}{0.49\linewidth}
		\centering
        \includegraphics[width=1\linewidth]{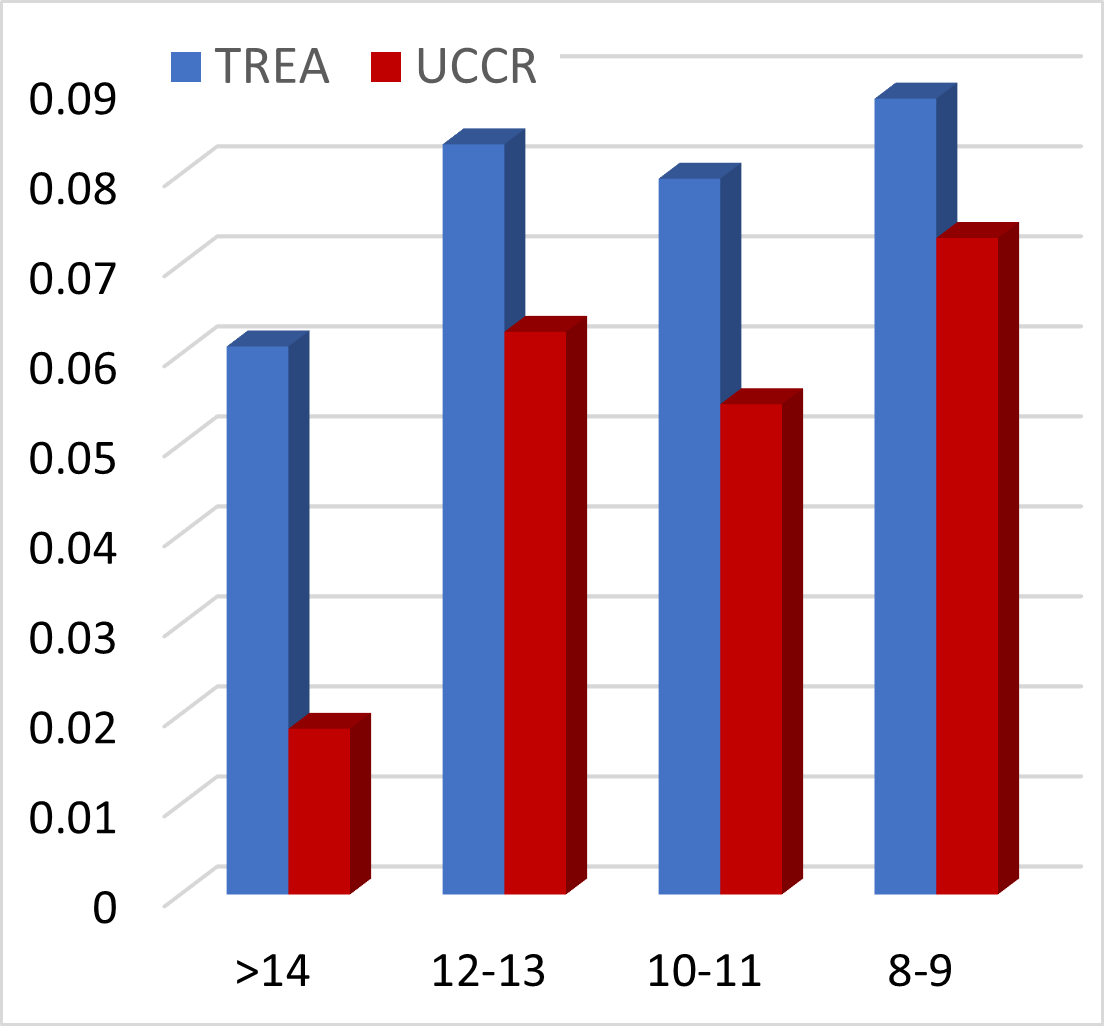}
        \caption{TG-ReDial}
	\end{subfigure}
	\caption{Evaluation results (R@50) of TREA and UCCR on data of different converstaion rounds.}
	\label{long}
\end{figure}

\subsection{Evaluation on Long Conversations}
We further evaluate TREA in long conversation scenarios. To the best of our knowledge, it is the first time to discuss this aspect of CRS. When the dialogue becomes longer and more knowledge information appears, if the relationships between knowledge pieces are not clarified, the model is not able to utilize the historical information effectively. We evaluate our TREA and a competitive baseline UCCR on data of different conversation rounds, measured by the metric Recall@50. The results in Figure \ref{long} shows that the performance of UCCR decreases sharply when the conversation rounds exceed 12 in ReDial and 14 in TG-ReDial. On the contrary, the performance of TREA fluctuates less as the number of conversation rounds increases. It indicates that the reasoning process of TREA can illuminate sophisticated relationships between historical entities for a better reference to the current situation, which further proves that nonlinear reasoning with historical hierarchy is vital to modeling user preference, especially when the conversation is long and the informativeness is great.
\section{Conclusion}
In this paper, we propose a novel tree-structure reasoning schema for CRS to clarify the sophisticated relationships between mentioned entities for accurate user modeling. In the constructed reasoning tree, each entity is connected to its cause which motivates the mention of the entity to provide a clear reference for the current recommendation. The generation module also interacts with the reasoning tree to extract relevant textual information. Extensive experimental results have shown that our approach outperforms several competitive baselines, especially in long conversation scenarios.

\section{Limitations}
The construction of the reasoning tree may be affected by the KG quality since the connection operations are variant with the KG structure. Hence the unsolved problem in Knowledge Graph such as incompleteness or noise could disturb the reasoning process. In the future, we will explore a solution to alleviate the influence of the side information.

\section*{Acknowledgements}
This work was supported in part by the National Natural Science Foundation of China under Grant No.62276110, No.62172039 and in part by the fund of Joint Laboratory of HUST and Pingan Property \& Casualty Research (HPL). The authors would also like to thank the anonymous reviewers for their comments on improving the quality of this paper. 
\bibliographystyle{acl_natbib}
\bibliography{custom}

\end{document}